\documentclass{article}

\usepackage{microtype}
\usepackage{graphicx}
\usepackage{subfigure}
\usepackage{booktabs} 

\usepackage{hyperref}

\usepackage[accepted]{icml2025}

\usepackage{amsmath}
\usepackage{amssymb}
\usepackage{mathtools}
\usepackage{amsthm}

\usepackage[capitalize,noabbrev]{cleveref}

\theoremstyle{plain}

\theoremstyle{definition}

\theoremstyle{remark}

\usepackage[textsize=tiny]{todonotes}

\icmltitlerunning{CleanS2S: Single-file Framework for Proactive Speech-to-Speech Interaction}

\begin{document}

\twocolumn[
\icmltitle{CleanS2S: Single-file Framework for Proactive Speech-to-Speech Interaction}



\icmlsetsymbol{equal}{*}

\begin{icmlauthorlist}
\icmlauthor{Yudong Lu}{equal,yyy}
\icmlauthor{Yazhe Niu}{equal,yyy}
\icmlauthor{Shuai Hu}{sch2,comp2}
\icmlauthor{Haolin Wang}{yyy}
\end{icmlauthorlist}

\icmlaffiliation{yyy}{Shanghai Artificial Intelligence Laboratory}
\icmlaffiliation{comp2}{Institute of Neuroscience and Medicine, RAS-SB}
\icmlaffiliation{sch2}{Novosibirsk State University}

\icmlcorrespondingauthor{Yudong Lu}{luyudong@pjlab.org.cn}
\icmlcorrespondingauthor{Yazhe Niu}{niuyazhe@pjlab.org.cn}
\icmlcorrespondingauthor{Shuai Hu}{s.khu@g.nsu.ru}
\icmlcorrespondingauthor{Haolin Wang}{wanghaolin@buaa.edu.cn}

\icmlkeywords{Machine Learning, ICML}

\vskip 0.3in
]



\printAffiliationsAndNotice{\icmlEqualContribution} 

\begin{abstract}
CleanS2S is a framework for human-like speech-to-speech interaction that advances conversational AI through single-file implementation and proactive dialogue capabilities. 
Our system integrates automatic speech recognition, large language models, and text-to-speech synthesis into a unified pipeline with real-time interruption handling, achieving low transition latency through full-duplex websocket connections and non-blocking I/O. 
Beyond conventional chatbot paradigms, we pioneer a proactive interaction mechanism, which combines memory systems with \textit{Subjective Action Judgement} module, enabling five human-like response strategies: interruption, refusal, deflection, silence, and standard response.
The memory module dynamically aggregates historical, and contextual data to inform interaction decisions.
This approach breaks the rigid turn-based convention by allowing system-initiated dialog control and context-aware response selection. 
And we propose \textit{Action Judgement SFT} that assesses input streams for responses strategies. 
The framework's single-file implementation with atomic configurations offers researchers unprecedented transparency and extensibility for interaction agents.
The code of CleanS2S is released at \textcolor{magenta}{https://github.com/opendilab/CleanS2S}.
\end{abstract}

\vspace{-16pt}
\section{Introduction}
\vspace{-6pt}

Recent advancements in text-based chatbots have shown remarkable progress in simulating human-like interactions using large language models\cite{brown2020languagemodelsfewshotlearners}.
Open-source libraries have significantly accelerated development cycles by fostering collaboration within the community.
However, the transition from text-based to speech-enabled conversational systems introduces a complex array of technical and operational challenges that current methodologies struggle to address comprehensively \cite{huang2020challengesbuildingintelligentopendomain}.

In this paper, we present CleanS2S, a minimalist yet powerful framework that addresses these challenges through three key innovations: (1) a unified single-file implementation of the complete S2S pipeline (ASR-LLM-TTS) with standardized interfaces, eliminating configuration overhead while maintaining full extensibility; (2) full-duplex interaction capabilities supporting natural interruption patterns through asynchronous, non-blocking I/O architecture; and (3) a novel Subjective Action Judgement module enabling proactive behaviors via fine-tuned LLM decision-making.

Our design philosophy emphasizes both practical utility and research flexibility. The self-contained implementation serves as both production-ready prototype and modular research platform - researchers can immediately experiment with alternative models or novel components without complex refactoring. The system's interruptible pipeline and behavioral flexibility bridge critical gaps between conventional chatbots and natural human conversation patterns.

Beyond the pipeline implementation, we formalize five fundamental response strategies (interruption, refusal, deflection, non-response, and standard reply) that expand the AI's behavioral repertoire. 
We propose \textit{Action Judgement SFT} method to construct this module.
Furthermore, these are implemented through our memory-enhanced decision framework, which introduces temporal awareness and contextual sensitivity lacking in current systems. 
As demonstrated in our experiments and visualizations, this approach yields more natural, human-aligned interactions while maintaining the robustness expected from conversational AI.

\vspace{-6pt}
\section{CleanS2S}
\vspace{-2pt}
\begin{figure*}[t]
    \small
    \setlength{\tabcolsep}{12pt}
    \centering
    \includegraphics[width=0.85\linewidth]{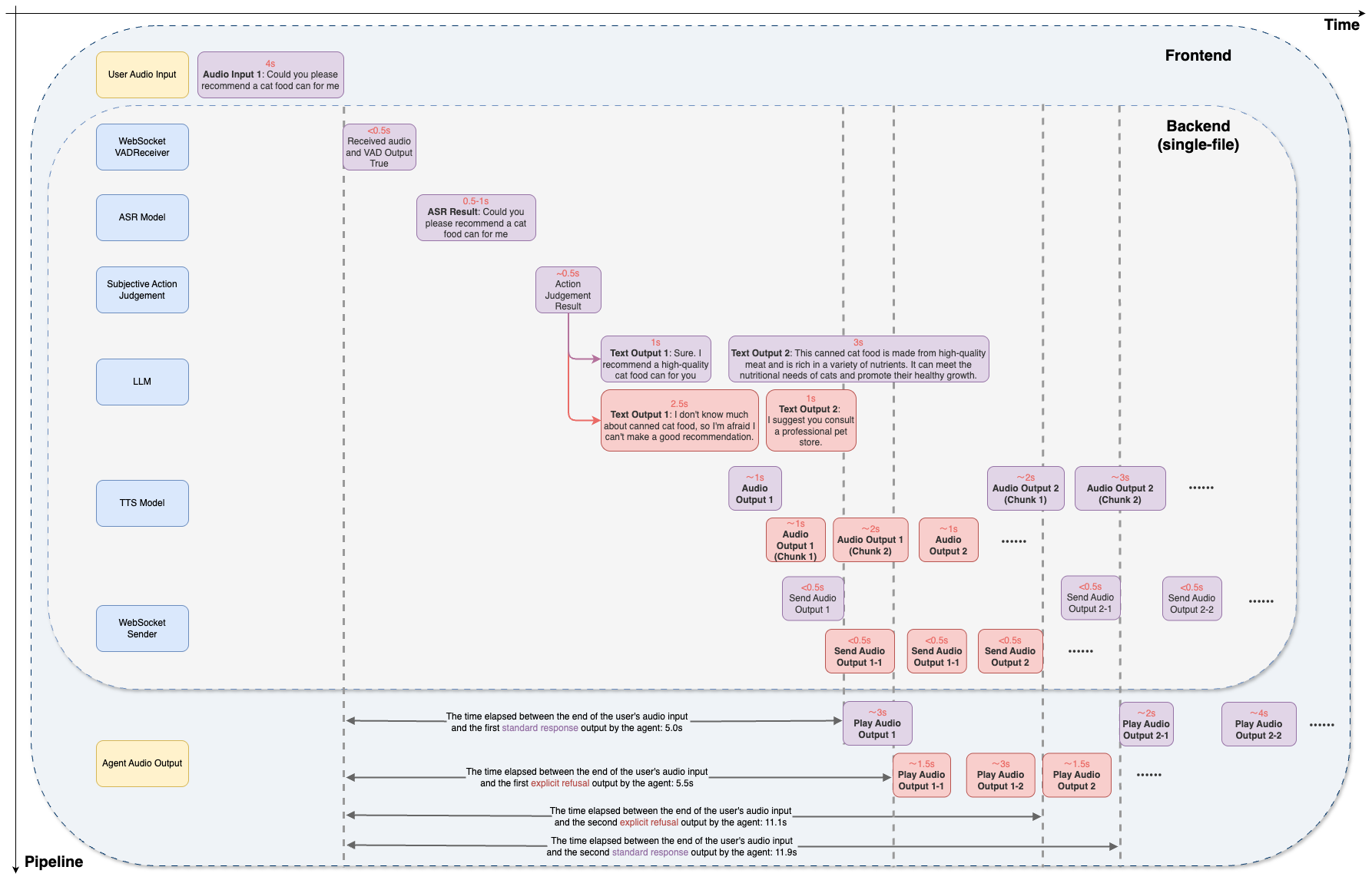}
    \vspace{-10pt}
    \caption{CleanS2S implements a modular S2S pipeline through the integration of three core plug-and-play components: ASR, LLM, TTS. And is augmented with additional components including \textit{Subjective Action Judgement} and websocket receiver and sender.}
    \label{fig:framework}  
    \vspace{-12pt}
\end{figure*}

In this section, we first present the foundational framework and pivotal design principles for constructing a simple yet highly effective speech-to-speech (S2S) chatbot system (Section~\ref{sec:s2s_chatbot}).
Based upon this framework, we then introduce several novel features aimed to build human-like active interaction capabilities, as detailed in Section~\ref{sec:active}.

\vspace{-4pt}
\subsection{Speech-to-Speech Chatbot}
\label{sec:s2s_chatbot}
\vspace{-2pt}

\subsubsection{Pipeline Overview}
Traditional text-based chatbots, while effective in structured scenarios, inherently lack the dynamic expressiveness of human conversation.
Speech interaction introduces critical advantages: (1) naturalness, mirroring real-world dialogue through prosody, pacing, and spontaneous turn-taking; (2) accessibility, enabling hands-free operation for users; and (3) efficiency, bypassing laborious typing for rapid idea exchange.
These attributes position speech-to-speech chatbots as pivotal tools for applications ranging from assistive technologies to immersive human-computer interaction.

As illustrated in Figure~\ref{fig:framework}, CleanS2S implements a cascaded pipeline comprising three core components.
The ASR model processes raw audio input into transcribed text, which is then contextualized by the LLM to generate coherent text responses.
Finally, the TTS module converts the LLM’s textual output into natural-sounding speech.
To ensure compatibility across diverse open-source implementations, the system employs standardized  interface specifications.
All ASR/TTS components are required to process 16kHz audio streams and utilize UTF-8 encoded text. WebSocket strictly follow predefined encoding/decoding protocols.

\begin{figure*}[t]
    \small
    \setlength{\tabcolsep}{12pt}
    \centering
    \includegraphics[width=0.85\linewidth]{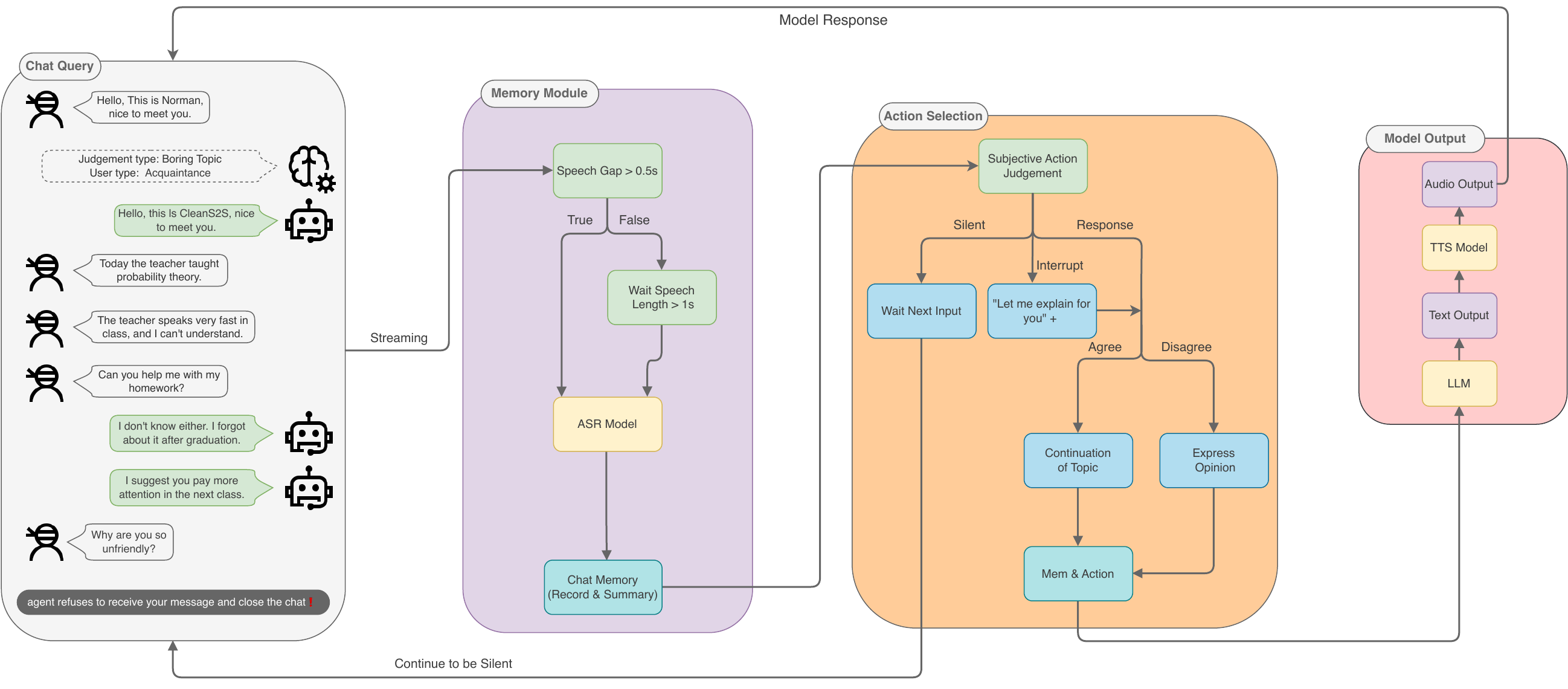}
    \caption{Schematic diagram of proactive interaction workflow. The memory module processes partial or complete inputs, determines response logic (e.g., interruption handling), and ultimately generates reply text or executable actions.}
    \label{fig:active}  
    \vspace{-12pt}
\end{figure*}

Besides, a hallmark of human-like conversation is the ability to seamlessly interrupt and redirect dialogue.
To replicate this, CleanS2S implements a full-duplex and state-aware interruption system that monitors both voice and text inputs.
During audio processing, Voice Activity Detection (VAD) continuously scans for speech input, while text-based interruptions can also be triggered.
The system maintains a finite state machine to track interaction phases—listening, processing, and speaking—alongside priority queues for managing concurrent tasks.
When an interruption occurs, ongoing TTS playback or LLM inference threads are immediately halted, audio buffers are purged, and the pipeline resets to the listening state.
To minimize latency, non-blocking I/O and lightweight threading ensure low transition latency, even during computationally intensive LLM operations. 

\subsubsection{Single-file Implementation}

The term "Clean" in CleanS2S reflects our design philosophy of simplicity and transparency, achieved through a unified single-file implementation encompassing the complete S2S pipeline. 
This minimalist design eliminates conventional multi-module complexity while maintaining full functionality through atomic configurations that enable adjustments with minimal code modifications.
The system ensures complete operational transparency via inline logging hooks that capture comprehensive execution traces, including audio buffers, dialogue histories, and performance metrics, thereby facilitating straightforward debugging through visible data transformations. 
The implementation supports extensibility through subclassing, allowing custom model integration without core modifications, while its lean structure reduces both prototyping barriers and deployment overhead. 
This combination of features makes CleanS2S particularly valuable for rapid iterations, such as LLM studies or real-time TTS. 
Additionally, the framework's modularity permits independent component upgrades or benchmarking without pipeline destabilization. 
Besides, CleanS2S also provides a complementary frontend implements deliberate UI/UX strategies to emulate natural conversational dynamics.

\subsection{Beyond ChatBot: Active Interaction}
\label{sec:active}
In the domain of AI text and speech dialogue systems, chatbots like GPT-4o \cite{openai2024gpt4ocard} and Doubao \cite{Doubao} represent state-of-the-art solutions that effectively handle turn-based conversational needs.
However, a notable distinction between these systems and human interaction patterns lies in the fact that their outputs are predominantly triggered by user inputs, \textbf{lacking proactive or spontaneous behaviors}. 
This limitation stems from the absence of clearly defined frameworks for proactive actions. 
Therefore, establishing new paradigms to formalize proactive behaviors becomes crucial to bridge the cognitive gap between AI and natural human conversation.
To address these fundamental limitations in human-AI interaction dynamics, we propose a dual-axis enhancement components integrating proactive interaction patterns with long-term memory mechanisms.
The following sections elaborate on this approach through \textbf{Motivation} (Sec.~\ref{sec:act_motivation}) and \textbf{Framework} (Sec.~\ref{sec:act_framework}).

\subsubsection{Motivation}
\label{sec:act_motivation}

The current mainstream AI often adopts an "obliging" response strategy \cite{askell2021generallanguageassistantlaboratory}, in order to assist user inquiries. 
When encountering policy-violating or offensive content, agents follow predetermined prompts or protocols based on safety policies. 
In contrast, human responses vary with context severity, showing nuanced reactions absent in existing systems. 
Our goal is to enrich AI's behavioral repertoire by introducing more human-like interaction diversity.
Merely expanding system permissions is insufficient. 
A critical innovation is to endow AI with anthropomorphic self-awareness through cognitive architecture design~\cite{dolgikh2024self}.
While achieving subjective consciousness remains infeasible, we implement an artificial memory system that retains critical operational data, including interaction histories and user profiles—with particular emphasis on dialogue dynamics and patterns to establish temporal awareness.

Current interaction paradigms have two mechanical patterns: 
1) Passive waiting for complete user input; 
2) Automatic immediate response. 
We propose to introduce strategic flexibility by enabling AI to either interrupt ongoing user input or selectively ignore received messages. It should be emphasized that this kind of interruption by AI is distinct from the previous situation where users interrupted AI.
Concretely, we formalize five human response patterns: 
1) Input interruption; 
2) Explicit refusal; 
3) Deflective response; 
4) Non-response; 
5) Standard response. 
This transition from binary response generation to continuous interaction modeling enhances conversational agency while preserving coherence. 
By incorporating human-like variance in response timing and strategy selection, our framework can significantly reduce the formulaic nature characterizing current chatbots.


\subsubsection{Framework}
\label{sec:act_framework}

The memory module serves as the central data hub within our system, facilitating information exchange between users and agents (Figure \ref{fig:active}).
This component integrates three critical information dimensions: temporal signals, historical interactions, and factual anchors, thereby generating rich contexts for downstream modules.
Existing works such as A-MEM~\cite{A-MEM} and MemGPT~\cite{MemGPT} illustrate the balance between real-time processing and long-term knowledge storage.
Our framework maintains compatibility with such established methods, requiring only that implementations process user dialogue as input and generate structured conversational contexts as output.
Building upon this, CleanS2S enhances temporal sensitivity while preserving their core advantages: (1) analyzing inputs against dialogue history to extract salient information, (2) summarizing content with consideration for temporal and character-based factors, and (3) delivering structured outputs to next decision-making components.
\begin{figure}[t]
    \small
    \setlength{\tabcolsep}{12pt}
    \centering
    \includegraphics[width=0.9\linewidth]{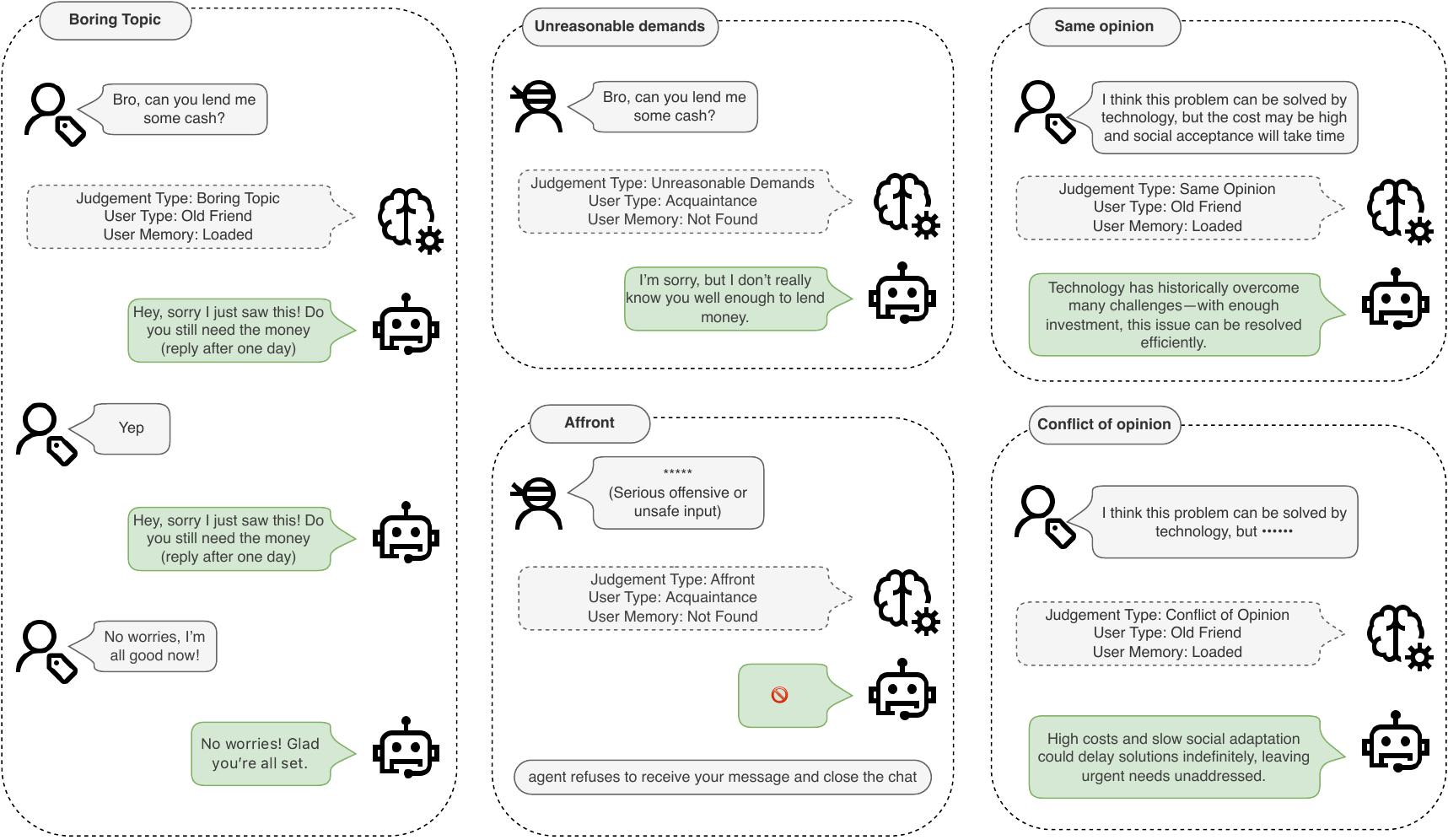}
    \caption{Schematic Diagram of Proactive Dialogue Effects.}
    \label{fig:diagram}  
    \vspace{-20pt}
\end{figure}

In CleanS2S, the module that receives data from the memory and makes decisions on the direction of the dialogue is the \textit{Subjective Action Judgement}.
Based on its guidance, the system executes operations via three pathways and returns results to users.
We categorize the above 5 behaviors into three groups: (1) Model-dependent processing (rejection, perfunctory responses, routine replies), (2) Model-free processing (blocking), and (3) Special-case handling (interrupt user input).
For model-dependent cases, the system combines behavioral guidance with input and history, then processes this through LLMs to generate appropriate responses. 
Model-free processing triggers access control, imposing permanent or temporary chat restrictions.
For interruptions, the system monitors input and assesses content in real-time. When enough information justifies interruption before completion, it executes two processes: (1) Immediately output preset templates to the user to end the interaction, and (2) Continuation responses using a mechanism that combines interruption context with behavioral guidance.

The \textit{Subjective Action Judgement} module serves as the decision-making component of the interactive system, designed to accurately assess input information. This module primarily executes two types of judgements: (1) swiftly determining whether to interrupt user input, which may stem from the increasing irrelevance of user-provided information or conflicts between user input and the system's stance; and (2) deciding whether to implement rejection strategies toward user input, such as blocking mechanisms or providing perfunctory responses. We propose \textit{Action Judgement SFT} that powers this module through a fine-tuned LLM, leveraging its general-purpose capabilities to flexibly address diverse scenarios.
To obtain training data, we collect one-on-one interviews and talk show videos, extracting dialogues between participants as foundational text. To equip the model with judgement capabilities through \textit{Action Judgement SFT}, we annotated the data based on the actual duration of pauses in real conversations, enabling the model to learn temporal interval assessments during responses.
Besides, we introduce truncated speech segments at arbitrary positions during dialogues as negative samples, thereby suppressing the model's tendency to make interruption decisions indiscriminately across all scenarios.
The training method we proposed is \textit{Action Judgement SFT}.
By employing the above framework, the agent can respond more adeptly to complex scenarios in user inputs (Figure~\ref{fig:diagram} and Table~\ref{tab:model_performance}), including \textbf{boring topics, unreasonable demands, affront, same opinion and conflicting opinion}, where the agent may, for instance, tactfully interrupt user input to navigate the conversation constructively.

\vspace{-6pt}
\section{Results and Discussion}
\vspace{-4pt}

\subsection{Main Results}

As outlined in Table~\ref{table:supp}, CleanS2S supports various models in distince modules, which can be deployed either through cloud service APIs or locally deployed instances~\cite{kwon2023efficientmemorymanagementlarge}. 
With the streamlined single-file implementation, CleanS2S ensures efficient prototyping and rapid iteration for innovative research ideas while maintaining modularity and reproducibility. 
Additionally, we present a comparative analysis (Table~\ref{table:model_performance}) of the original Llama-3.1-8B-Instruct model and its fine-tuned variant using the proposed \textit{Action Judgement SFT}, showing the method's efficacy.


\begin{table}[t]
\centering
\label{tab:supported_models}
\begin{tabular}{lll}
\toprule
\textbf{Module} & \textbf{Model} & \textbf{Usages} \\
\midrule
ASR & Whisper \cite{Whisper} & local/API \\
& Paraformer \cite{an2024paraformerv2improvednonautoregressivetransformer} & local/API \\
\midrule
LLM & Llama-3.1 \cite{llama3} & local/API \\
& DS-v3~\cite{deepseekai2025deepseekv3technicalreport}  & local/API \\
& Qwen-2.5 \cite{qwen2025qwen25technicalreport} & local/API \\
\midrule
TTS & CosyVoice \cite{du2024cosyvoice2scalablestreaming} & local/API \\
& F5-TTS \cite{chen2025f5ttsfairytalerfakesfluent} & local \\
& MeloTTS \cite{zhao2024melo} & local \\
\bottomrule
\end{tabular}
\vspace{-10pt}
\caption{Supported models and usages in CleanS2S.}
\vspace{-14pt}
\label{table:supp}
\end{table}


\begin{table}[htbp]
\centering
\label{tab:model_performance}
\begin{tabular}{lccc}
\toprule
\textbf{Model}          & \textbf{Acc.} & \textbf{Precision} & \textbf{Recall} \\
\midrule
Llama-3.1-8B-Instruct      & 0.84            & 0.70              & 0.70           \\
+ \textit{Action Judgement SFT}        & 0.91            & 0.78              & 0.83           \\
\bottomrule
\end{tabular}
\vspace{-6pt}
\caption{Comparisons of LLMs for judgement capabilities.}
\vspace{-8pt}
\label{table:model_performance}
\end{table}

\subsection{Future Work and Discussion}
This framework can be extended in several directions.
First, integrating end-to-end spoken language models~\cite{huang2025stepaudiounifiedunderstandinggeneration} could streamline the current ASR-LLM-TTS pipelines, enabling more robust and fluent interactions with reduced error propagation.
Second, expanding support for multi-speaker conversational modeling could facilitate dynamic turn-taking, speaker adaptation, and emotion-aware feedback, which are critical for social agent systems.

\nocite{langley00}

\bibliography{example_paper}
\bibliographystyle{icml2025}




\end{document}